\documentclass[letterpaper, 10 pt, conference]{ieeeconf}  

\usepackage{epsfig}
\usepackage{epstopdf}
\usepackage{graphics}
\usepackage{epstopdf}
\usepackage{upgreek}
\usepackage{amssymb}
\usepackage{amsfonts}
\usepackage{amsmath}
\usepackage{refstyle}
\usepackage{xcolor}
\usepackage{subfigure}
\usepackage{todonotes}
\usepackage{mathtools}
\usepackage{algorithm}
\usepackage{algpseudocode}
\usepackage{cite}
\usepackage{balance}

\IEEEoverridecommandlockouts                              

\title{\LARGE \bf
Spotlight-based 3D Instrument Guidance for Retinal Surgery}

\author{Mingchuan Zhou$^{1,2}$,~\IEEEmembership{Member,~IEEE}, Jiahao Wu$^{1,3}$,~\IEEEmembership{Student Member,~IEEE}, Ali Ebrahimi$^{1}$,\\~\IEEEmembership{Student Member,~IEEE},
Niravkumar Patel$^{1}$,~\IEEEmembership{Member,~IEEE},
 Changyan He$^{1}$,~\IEEEmembership{Student Member,~IEEE}, Peter\\ Gehlbach$^{4}$,~\IEEEmembership{ Member,~IEEE}, Russell H. Taylor$^{1}$,~\IEEEmembership{Life Member,~IEEE}, Alois Knoll$^{2}$,~\IEEEmembership{Senior Member,~IEEE},\\ M. Ali Nasseri$^{5}$,~\IEEEmembership{Member,~IEEE}, and Iulian Iordachita$^{1}$,~\IEEEmembership{Senior Member,~IEEE} 
\thanks{
This work was supported by U.S. National Institutes of Health under grants number 1R01EB023943-01 and 1R01 EB025883-01A1. The work was  supported by Research to Prevent Blindness, New York, New York, USA, and gifts by the J. Willard and Alice S. Marriott Foundation, the Gale Trust, Mr. Herb Ehlers, Mr. Bill Wilbur, Mr. and Mrs. Rajandre Shaw, Ms. Helen Nassif, Ms Mary Ellen Keck, Don and Maggie Feiner, and Mr. Ronald Stiff. The work was also partly supported by TUM-GS internationalization funding.
 }
\thanks{
        $^{1}$ M. Zhou, J. Wu, A. Ebrahimi, N. Patel, C. He, R. Taylor, and I. Iordachita are with Department of Mechanical Engineering and Laboratory for Computational Sensing and Robotics at the Johns Hopkins University, Baltimore,
MD 21218 USA. \{{\tt\small mzhou28, jwu151, aebrahi5, npatel89, changyanhe, iordachita\} @jhu.edu}
		}		
\thanks{$^{2}$M. Zhou and A. Knoll are with Chair of Robotics, Artificial Intelligence and Real-time Systems, Technische Universit\"{a}t M\"{u}nchen, M\"{u}nchen 85748 Germany. \{{\tt\small zhoum, knoll\} @tum.de}
        }
\thanks{$^{3}$J. Wu is with the T Stone Robotics Institute, the Department of Mechanical and Automation Engineering, The Chinese University of Hong Kong, HKSAR, China. {\tt\small jhwu@mae.cuhk.edu.hk}
        }%
\thanks{$^{4}$P. Gehlbach is with the Wilmer Eye Institute, Johns Hopkins Hospital, Baltimore, MD 21287 USA. {\tt\small pgelbach@jhmi.edu}
        }%
\thanks{$^{5}$M. Nasseri is with Augenklinik und Poliklinik, Klinikum rechts der Isar der Technische Universit\"{a}t M\"{u}nchen, M\"{u}nchen 81675 Germany.
{\tt\small ali.nasseri@mri.tum.de}
        }%
}

\DeclarePairedDelimiterX{\norm}[1]{\lVert}{\rVert}{#1}
\begin{document}

\maketitle
\thispagestyle{empty}

\pagestyle{empty}

\begin{abstract}
Retinal surgery is a complex activity that can be challenging for a surgeon to perform effectively and safely. Image guided robot-assisted surgery is one of the promising solutions that bring significant surgical enhancement in treatment outcome and reduce the physical limitations of human surgeons. In this paper, we demonstrate a novel method for 3D guidance of the instrument based on the projection of spotlight in the single microscope images. The spotlight projection mechanism is firstly analyzed and modeled with a projection on both a plane and a sphere surface. To test the feasibility of the proposed method, a light fiber is integrated into the  instrument which is driven by the Steady-Hand Eye Robot (SHER). The spot of light is segmented and tracked on a phantom retina using the proposed algorithm. The static calibration and dynamic test results both show that the proposed method can easily archive 0.5 mm of tip-to-surface distance which is within the clinically acceptable accuracy for intraocular visual guidance.
\end{abstract}

\section{Introduction}
\label{intro}
\begin{figure}[htbp]
\begin{minipage}[t]{0.95\linewidth}
    \centering
    \includegraphics[trim=0 0 0 0,clip,width=1.0\textwidth]{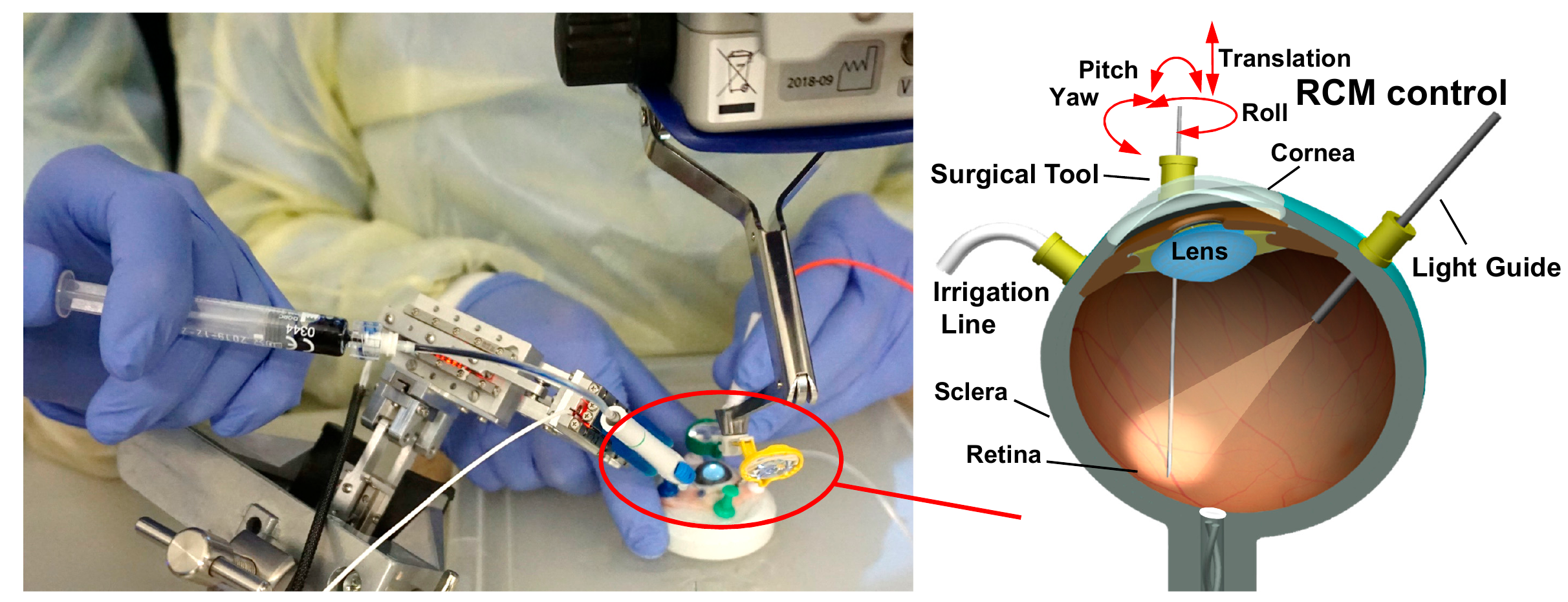}
    \parbox{9cm}{\small \hspace{2.3 cm}(a)\hspace{3.2 cm}(b)}
\end{minipage}
\caption{(a) The robot-assisted retinal surgery setup on the ex-vivo pig eye using iRAM!S robot~\cite{nasseri2013kinematics}. The robot hold the instrument instead of the surgeon's hand. (b) The cross-section view of surgical setup. The needle movement is restricted by RCM control where two translation motions are constrained.}
\label{fig:intro}
\end{figure}

Robot-assisted surgery (RAS) setups are known as the solution for reducing the work intensity, increasing the surgical outcomes and prolonging the viable service time of experienced surgeons in ophthalmic surgery. The introduction of autonomy into RAS can potentially help the surgeon perform the surgery with better outcomes and higher efficiency~\cite{moustris2011evolution}. Retinal surgery contains a complex workflow and delicate tissue manipulation (see Fig.~\ref{fig:intro}), which needs critical surgical skills and considerations. In 2019, there have been more than 342 million patients having retinal disease, mainly with age-related
macular degeneration (196 million) and diabetic retinopathy (146 million)~\cite{who}. Many of these patients lack access to proper and timely treatment and increase their chances of blindness~\cite{chou2013age}.

In September 2016, surgeons at Oxford's John Radcliffe Hospital performed the world's first robot-assisted eye surgery. The eye surgical robot named Robotic Retinal Dissection Device (R2D2) with 10 $\upmu$m position resolution was used in the clinical trials for subretinal injection~\cite{edwards2018first}. They used the microscope-integrated Optical Coherence Tomography (MI-OCT) (RESCAN 700, Carl Zeiss Meditec AG., Germany) to enhance the visual feedback during needle insertion beneath the retina. The OCT imaging modality provides a very suitable resolution for retinal surgery with around 2 $\upmu$m in depth but with image range limited to around 2 mm~\cite{zhou2019towards}. Zhou et al.~\cite{zhou2018icra,zhou2019towards,zhou20196dof} performed the preliminary research to show the feasibility of needle manipulation using an eye surgical robot with OCT images for navigation. However, the contradictory resolution of the image and image range results in less suitable for large range guidance of instrument movement. The instrument movement range depends on different application and cloud be treated as a volume of 10 mm$\times$10 mm$\times$5 mm based on the microscope view for some typical retinal surgeries, e.g. vessel tracking, subretinal injection, membrane peeling.

To navigate the instrument intraocular in 3D with a large range, Probst et al. proposed~\cite{probst2018automatic} the stereo-microscope vision system which uses deep learning to reconstruct the retina surface and localize the needle tip. The benefit of this method is that it will not introduce any other additional instrument inside the eye, while the drawback is that the deep learning method requires a big amount of data with annotation with different surgical tools. Different from this purely passive stereo-microscope vision system which can be influenced by the illumination, in this paper we proposed a proactive method using a spotlight source to navigate instrument 3D in retinal surgery. The contributions of this paper are listed as follows,
\begin{itemize}
\item  The theoretical model of the spotlight projection is analyzed for planar and spherical surfaces;
\item  An algorithm is developed for 3D navigating the instrument with spotlight projection;
\item  The calibration and dynamic tests are performed with Steady-Hand Eye Robot (SHER) to verify the feasibility and performance of the proposed method.
\end{itemize}

The remainder of the paper is organized as follows: in the next section, we briefly present related work. The proposed method is described in Section~\ref{sec:method}. In Section~\ref{sec:experiment}, the performance of the proposed method is evaluated and discussed. Finally, Section~\ref{sec:conclusion} concludes this paper and presents future work.

\section{Related work}
\label{sec:Relatedwork}
There are currently some possible solutions to navigate the instrument 3D intraocular. With the benefit of suitable resolution and radiationless imaging mechanism, the OCT imaging modality now is popular not only in the retina diagnostics but also in the intraoperative operation to enhance the visual feedback for the surgeon~\cite{roodaki2019real,tian2020toward,sandhu2020automated}. The research on instrument localization in OCT images has attracted the attention of researchers. Zhou et al.~\cite{zhou2017needle} introduced a fully conventional neural network to segment the needle in volumetric OCT images when the needle is above the tissue. Weiss et al.~\cite{weiss2018fast} introduced a method to estimate the 5 degree of freedom (DoF) needle pose for tool navigation during subretinal injection. Gessert et al.~\cite{gessert2018deep} introduced a 3D convolutional neural network to directly estimate a marker's 6D pose from the OCT volume. They used a marker with obvious geometrical features instead of a surgical instrument to simplify the problem. However, the OCT has limited image range in depth direction which makes it not suitable for large range guidance of instrument movement.  

Many studies have been carried out with significant progress in the instrument tracking and localization using microscopic image with a large movement range~\cite{li2014instrument,rieke2015surgical,sznitman2012data}. These works achieved satisfactory results using either color-based or geometry-based features. However, due to the limitation of using purely 2D microscopic image, these methods can not provide additional depth information to navigate the instrument in 3D. To  localize the instrument~\cite{yang2018techniques},  Yang et al. used a cone beam with structured-light reconstruction to estimate surface in the coordinate system of custom-built optical tracking system (ASAP) with additional stage, which is very related to our research. After surface reconstruction, the needle tip to surface distance is estimated in the coordinate system of ASAP~\cite{yang2016comparative}. Differently, our proposed method  integrated the beam into the instrument, thus directly estimate the tool tip to surface distance in real-time without complications of surface reconstruction. 

Probst et al. proposed~\cite{probst2018automatic} the stereo-microscope vision system which uses deep learning to reconstruct the retina surface and localize the needle tip. This method can obtain accuracy with around 0.1 mm in 3D over a large range. The drawback is that the deep learning method requires a large amount of data for annotation with different surgical tools and while other passive vision systems are influenced by  variations in illumination. To overcome this problem, we propose a novel method in which a proactive spotlight integrated into the instrument is used to navigate the instrument in 3D via a single microscope image. The projection of spotlight from a single image can be used to locate the surgical target on the retinal surface and then infer the distance to the retinal surface. An advantage of the proactive light is that projection detection is less influenced by variations in illumination.

\section{Method}
\label{sec:method}

\subsection{Distance estimation using spotlight projection }
\begin{figure}[htbp]
\centering
\begin{minipage}[t]{0.8\linewidth}
    \centering
    \includegraphics[width=1\columnwidth]{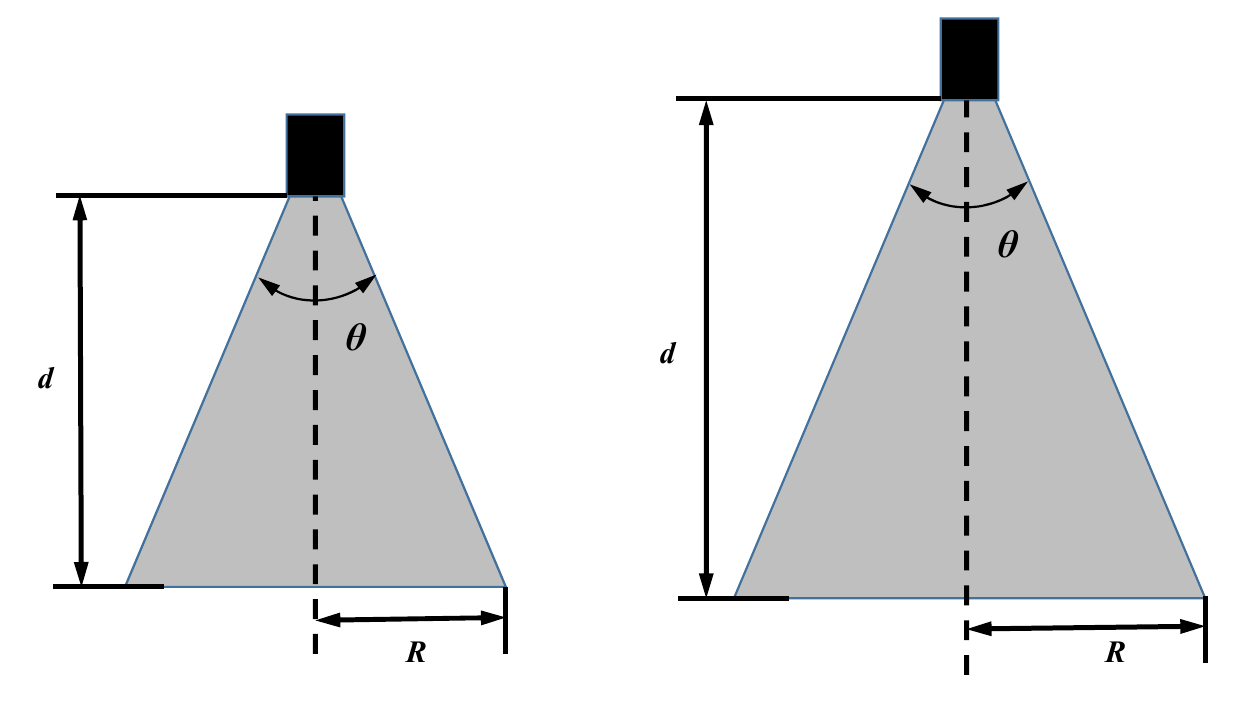}
\end{minipage}
\caption{The basic mechanism of spotlight. The distance $d$ can be inferred from the projected pattern.}
\label{fig:beamvertical}
\end{figure}

The spotlight contains the pattern itself with a very simple structure. The lens at the end of spotlight beam is used to concentrate the light. The light after being concentrated is a cone which intersects the surface as a circle on the surface when the light beam is targeted perpendicular to the flat surface, as shown in Fig.~\ref{fig:beamvertical}. The distance $d$ between the spotlight end-tip and surface is calculated as follows,

\begin{equation}
\label{equa:basic2}
d=kR+b
\end{equation}
where $b$ is the diameter of the circular shape when spotlight has zero distance to the plane, $k$ is related to the property of the lens, which equals $1/tan(\frac{\theta}{2})$, $\theta$ denotes the angle of the cone.

\begin{figure}[htbp]
\centering
\begin{minipage}[t]{0.99\linewidth}
    \centering
    \includegraphics[width=1\columnwidth]{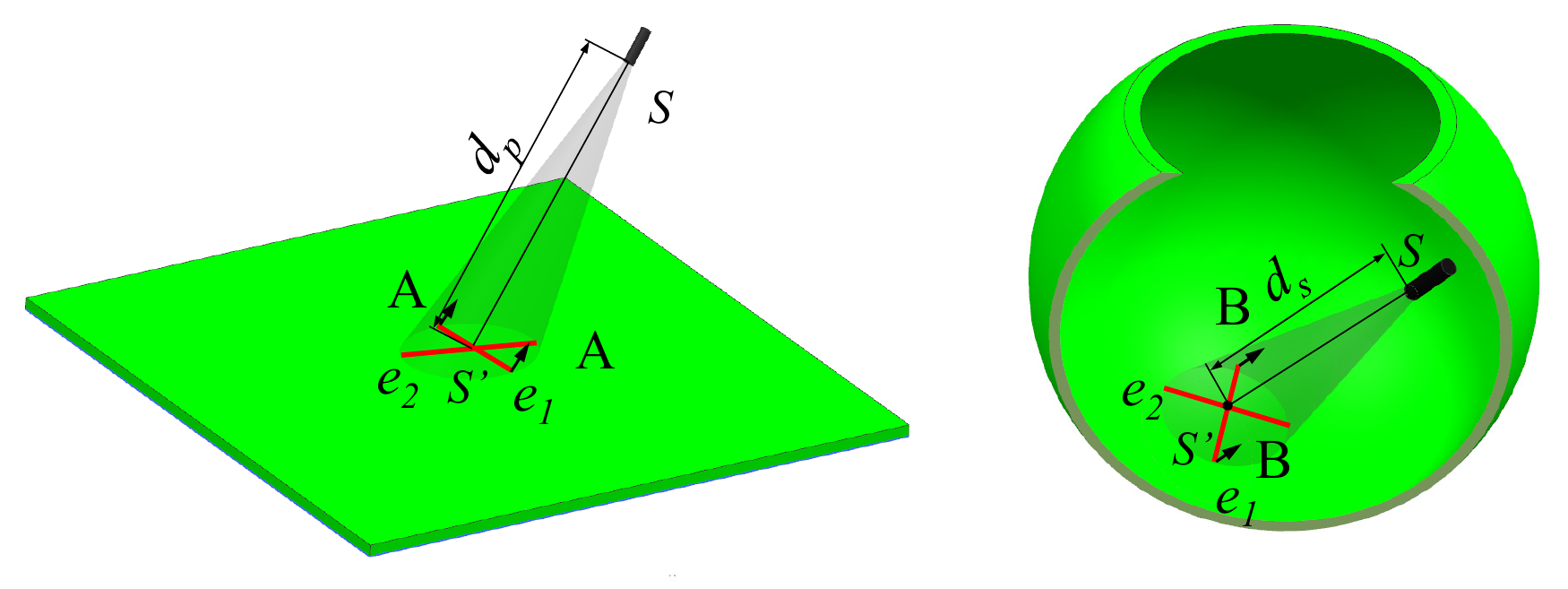}
        \parbox{6cm}{\small \hspace{0.9cm}(a)\hspace{3.6cm}(b)}
\end{minipage}
\begin{minipage}[t]{0.99\linewidth}
    \centering
    \includegraphics[width=1\columnwidth]{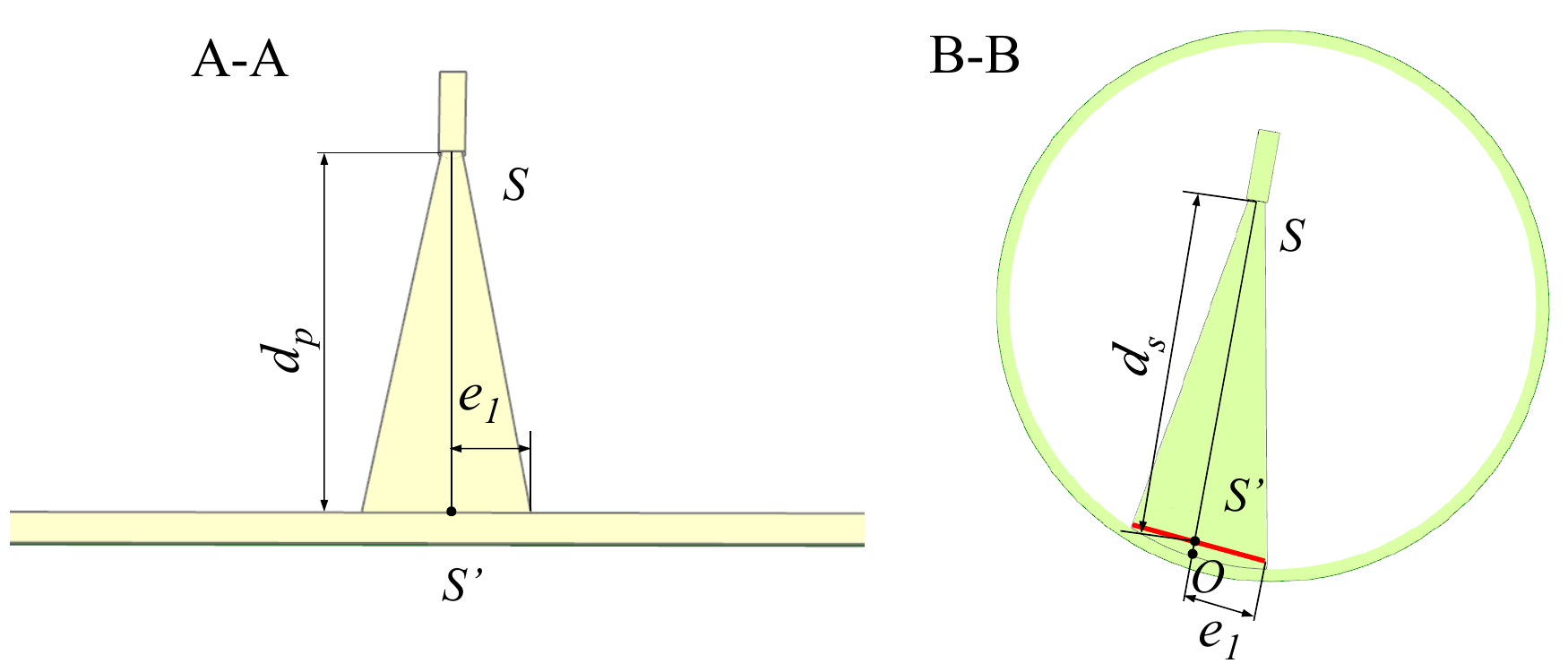}
        \parbox{6cm}{\small \hspace{0.9cm}(c)\hspace{3.6cm}(d)}
\end{minipage}
\caption{(a) The spotlight has a non-perpendicular angle with the plane surface. $SS'$ denotes the generatrix of cone where $S$ is the spotlight end-tip and $S'$ is the intersection point of the generatrix and plane. $e_1$ and $e_2$ denote the short axis and long axis of the ellipse, respectively. (b) The spotlight inside a eyeball sphere. (c) The cross-section of A-A plane in (a). (d) The cross-section of B-B plane in (b). $O$ is the the intersection point of the generatrix and sphere.}
\label{fig:diffsurface}
\end{figure}

When the spotlight is placed with a non-zero angle to plane surface, as shown in Fig.~\ref{fig:diffsurface}(a), the projection of the pattern will be an ellipse. Based on the cross-section A-A shown as in Fig.~\ref{fig:diffsurface}(c), we can obtain the following equation,
\begin{equation}
\label{equa:basic3}
d_p=ke_1+b
\end{equation}
where $d_p$ is the distance between the spotlight end-tip and plane surface. $e_1$ is the short axis of the ellipse.

During the retinal surgery, the spotlight is projected on the intraocular surface which can be considered as a sphere with diameter of 22-23 mm~\cite{bekerman2014variations}. Due to the projected pattern is controlled within around 2 mm light spot, which is significantly smaller than the eyeball diameter, shown in Fig.~\ref{fig:diffsurface} (b) and (d), the projected area on the sphere can be treated as a small plane to simplify the calculation. The distance can be calculated as,
\begin{equation}
\label{equa:basic4}
d_s=SO=SS'+S'O=ke_1+b+S'O
\end{equation}
where $d_p$ is the distance between the spotlight end-tip and plane surface. $SO$ can be considered as the radius r, when the distance is small. $S'O$ is related to the value of $e_1$ which can be calculated as,
\begin{equation}
\label{equa:basic5}
S'O=r-\sqrt{r^2-e_1^2}
\end{equation}
where $r$ is the radius of the eyeball in the cross section of B-B. Combining Eq.~\ref{equa:basic4} and Eq.~\ref{equa:basic5}, \begin{equation}
\label{equa:basic6}
d_s=ke_1+b+r-\sqrt{r^2-e_1^2}
\end{equation}

The microscope for retinal surgery is considered as a pinhole camera model. If the microscope and plane are parallel to each other, the detected ellipse short axis $e_1'$ in the microscope image has the proportional relation with the $e_1'$,
\begin{equation}
\label{equa:basic7}
d_p=\delta ke_1'+b
\end{equation}
where $\delta$ is the ratio factor.

\begin{figure}[htbp]
\centering
\begin{minipage}[t]{0.95\linewidth}
    \centering
    \includegraphics[width=1\columnwidth]{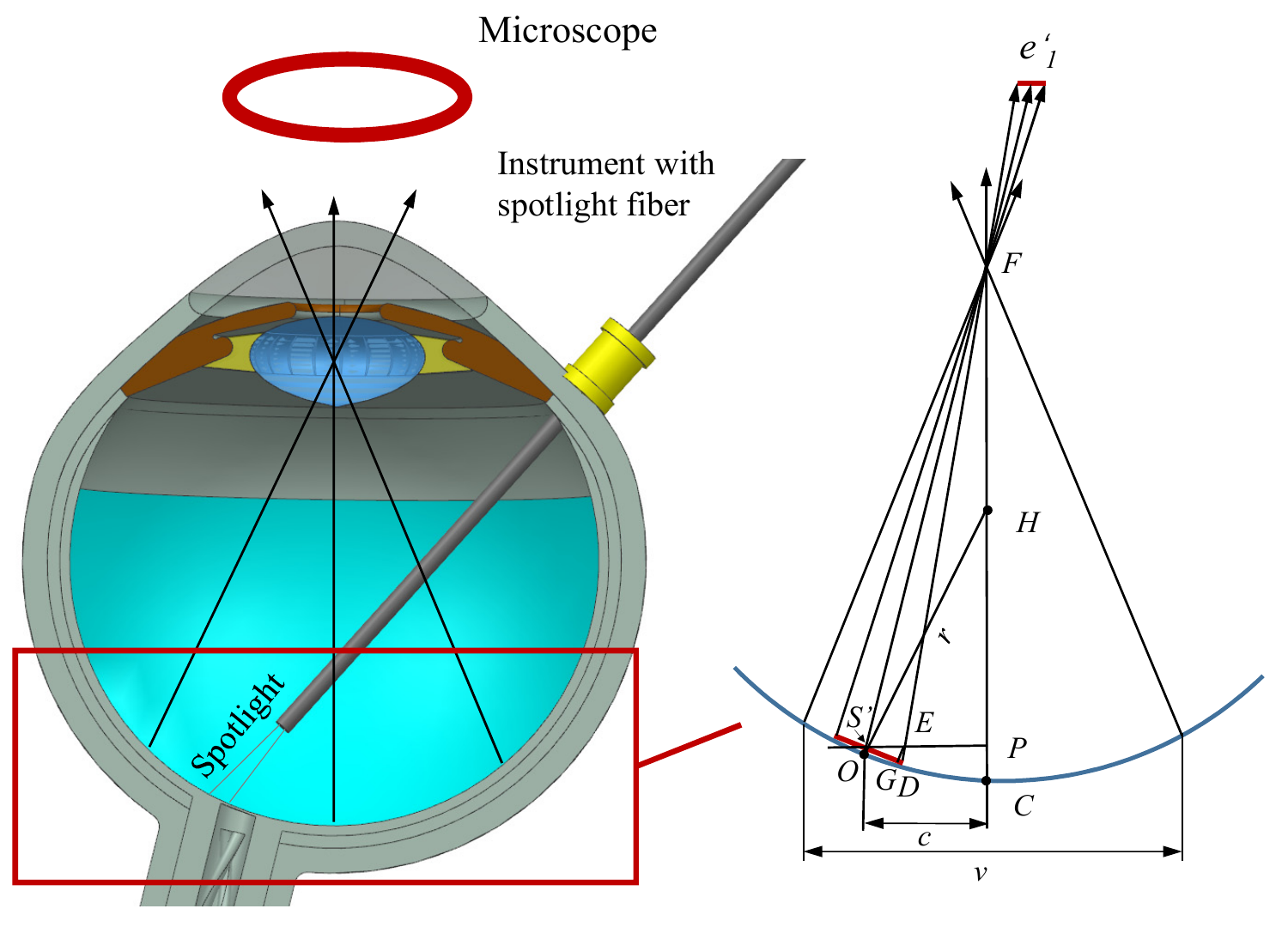}
\end{minipage}
\caption{The spotlight projected on the sphere. $FC$ denotes the center of microscope. $S'P$ is perpendicular with $FC$. $S'D$ denotes the length of $e_1$. $E$ is the intersection point between $S'P$ and $FD$. $EG$ is perpendicular with $OD$. $c$ denotes the distance between point $O$ and line $FC$. $v$ denotes the view size of microscope intraocular. F lies on the eyeball circle center.}
\label{fig:pinehole}
\end{figure}

However, when the spot light pattern is projected on the sphere, as shown in Fig.~\ref{fig:pinehole}, then $e_1$ will be influenced by other factor which can be calculated as follows,
\begin{equation}
\label{equa:basic8}
\begin{split}
e_1&=S'E cos\angle ES'D+DE cos\angle S'DE\\
&=\delta e_1' cos\angle ES'D+DE cos\angle S'DE
\end{split}
\end{equation}
where $\angle ES'D$ is the angle between the tangent line of the circle and the line of $S'E$. Due to the fact that the view size of microscope $v$ is within 10 mm typically, it can be inferred that $DE<S'E$. The minimum of $\angle S'DE$ appears when the $O$ locates at the edge of the view area.  $\angle S'DE$ can be calculated,

\begin{equation}
\label{equa:basic9}
\begin{split}
\angle S'DE & \approx atan(\frac{S'F}{S'D}) \approx atan(\frac{DF}{S'D})\\
&=atan(\frac{FPcos(asin(\frac{c}{r})/2)}{S'D})\\
&=atan(\frac{(r+\sqrt{r^2-c^2})cos(asin(\frac{c}{r})/2)}{S'D})
\end{split}
\end{equation}

Based on the above equations and  $v$ which is within 10 mm, the minimum value of $\angle S'DE$ is more than 87.5$^\circ$. Therefore, value of $DEcos \angle S'DE$ in Eq.~\ref{equa:basic8} is much smaller than $\delta S'E \angle ES'D$, which Eq.~\ref{equa:basic8} can be rewritten as,
\begin{equation}
\label{equa:basic10}
e_1 \approx \delta e_1'cos \angle ES'D=\delta e_1'cos(atan\frac{c}{\sqrt{r^2-c^2}})
\end{equation}

Combining Eq.~\ref{equa:basic6}, we can obtain,
\begin{equation}
\label{equa:basic10}
d=k\delta
cos(atan\frac{c}{\sqrt{r^2-c^2}})e_1'+b+r-\sqrt{r^2-e_1^2}
\end{equation}

\subsection{Image algorithm design}

\begin{figure}[htbp]
\centering
\begin{minipage}[t]{0.95\linewidth}
    \centering
    \includegraphics[width=1\columnwidth]{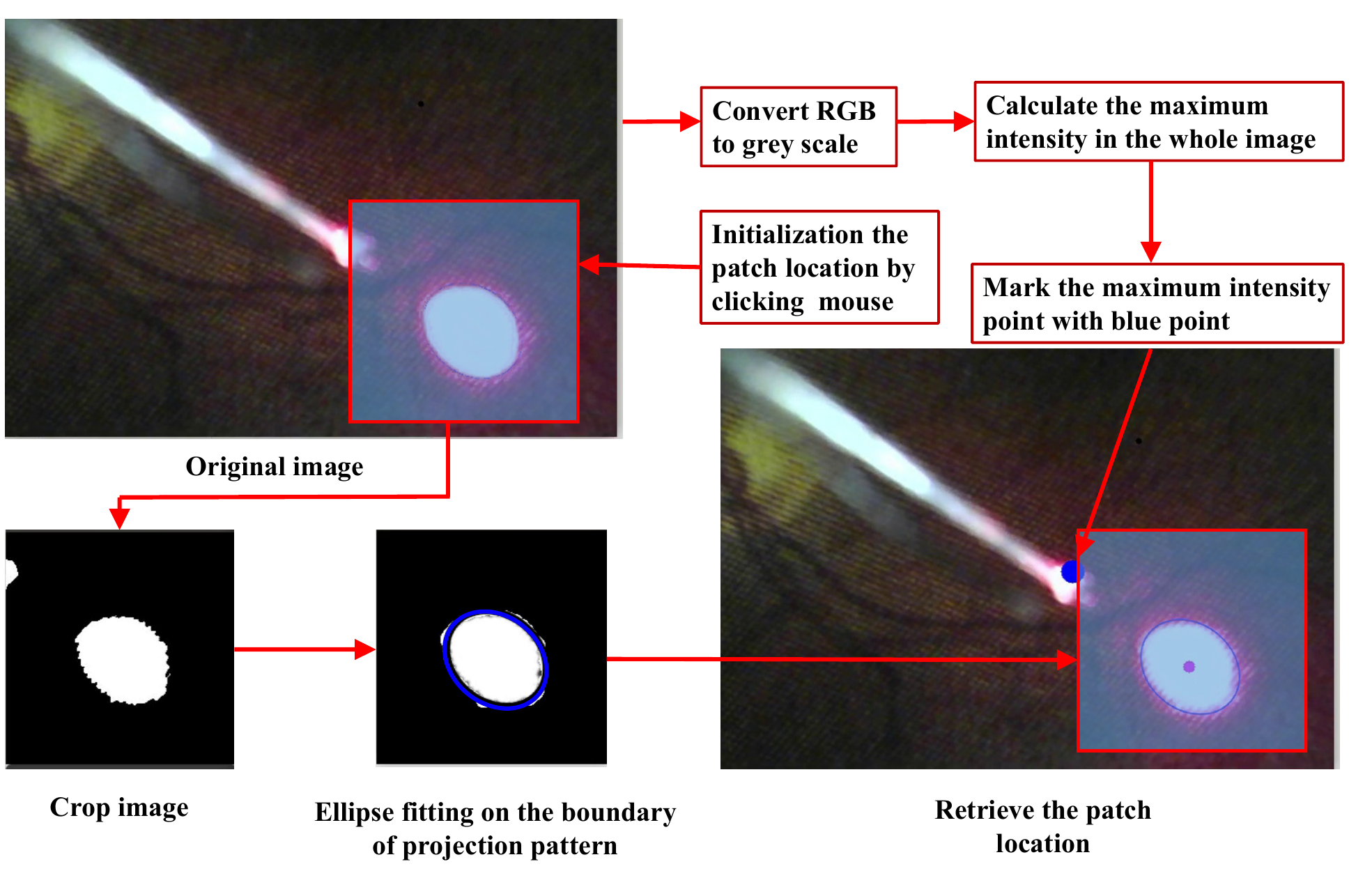}
\end{minipage}
\caption{The framework of projection pattern tracking. Spotlight projection is fitted with ellipse and tracked in every frame.}
\label{fig:designAlg}
\end{figure}

A framework is proposed to track the projection pattern on the surface from the spotlight, shown as in Fig.~\ref{fig:designAlg}. The input image is transferred into the gray image. In order to select the small projection pattern out of the whole image for reducing the unnecessary noise from background, the square area with a width size of $s$ is used to crop the gray image into a patch image. Since the spotlight projection pattern usually has brighter intensity compared to the rest of the background in the patch image, a fixed threshold (threshold intensity is 200 for 8 bits image) is used to segment the image into the binary image. We eliminate the noise inside the binary image, by applying a median filter followed by a Gaussian filter. To further assure the projection pattern in the patch image, the connected component labeling method is used to output the maximum number of the connected components, meanwhile, this maximum number also needs to be more than $T_s$. Based on the projection pattern binary image, an ellipse fitting is applied to the boundary of the projection pattern in the patch image. we constrain the ellipse's minor axis to a value lower than $m_e$,  where $m_e$ is defined based on the pre-defined range value of the ellipse's minor axis. The detailed algorithm description is shown in Algorithm~\ref{detect-needle}. Algorithm~\ref{patchcheck} explains the PatchCheck($M$) function in Algorithm~\ref{detect-needle}, which is to make sure the boundary of patch image is inside the original image.

\begin{algorithm}[!htb]
        \caption{Spotlight projection pattern tracking}
        \label{detect-needle}
         \textbf{INPUT:} $I_{RGB}$ - Input RGB image from microscope image\\
        \textbf{OUTPUT:} $E\{a_{min}, e_x, e_y\}$ - Output of parameters for the ellipse\\
       $RGB2GREY()\leftarrow$ Transfer RGB image to grey image\\
       $PatchCheck()\leftarrow$ Check whether path image boundary within the original image or not\\
       $Crop()\leftarrow$ Crop the image\\
       $Filter()\leftarrow$ Filter the image with a median filter and a Gaussian  filter\\
       $GREY2BW()\leftarrow$ Transfer grey image to binary image\\
       $BwLabel()\leftarrow$ Label the image with connected component method\\
       $EllipseFit()\leftarrow$ Find the ellipse boundary in the image
       \begin{algorithmic}[1]
            \Procedure{TrackingPattern}{$I_{RGB}$}
            \State $E\{a_{min}, e_x, e_y\} = \{\}$, $maxIndex = 0$, $vote = 0$
            \While {$I_{RGB}$ != $\varnothing$}
            \If {$Mouse\_left\_click$ == $true$}
            \State {$M$ = $Mouse\_cursor\_location$}
            \EndIf
            \If {$M\{m_x, m_y\}$ != $\varnothing$}
            \State $I_{GREY}=RGB2GREY(I_{RGB})$
            \If {$imcount==0$}
            \State {$M_c\{m_{cx}, m_{cy}\}$ = $PatchCheck$($M$)}
            \EndIf
            \State {$M_c\{m_{cx}, m_{cy}\}$ = $PatchCheck$($M$)}
            \State {$P=$$Crop$$(I_{GREY},[m_{cx}-p,m_{cy}-p,2p,2p])$}
            \State {$P$ = $Filter$($P$)}
            \State {$B$ = $GREY2BW$($P$)}
            \State {$L$ = $BwLabel$($B$)}
            \State {$E$ = $EllipseFit$($L$)}

            \If {$a_{min}<m_e$}
            \State Return $E\{a_{min}, e_x, e_y\}$
            \State {$M\{m_{x}=e_x, m_{y}=e_y\}$}
            \Else
            \State {$E\{a_{min}, e_x, e_y\}=\varnothing$}
            \EndIf
            \Else
            \State {$E\{a_{min}, e_x, e_y\}=\varnothing$}
            \EndIf
            \State Return $E\{a_{min}, e_x, e_y\}$
            \EndWhile
            \EndProcedure
        \end{algorithmic}
\end{algorithm}

\begin{algorithm}[!htb]
        \caption{Check patch boundary of the image}
        \label{patchcheck}
         \textbf{INPUT:} $M\{m_x, m_y\}$, $I_{GREY}$ - Input RGB image from microscope image\\
        \textbf{OUTPUT:} $M_c\{m_{cx}, m_{cy}\}$ - Output of parameters for the ellipse\\
        $SizeOf()\leftarrow$ Return the size of the image
        \begin{algorithmic}
            \Function{PatchCheck}{$M\{m_x, m_y\}$, $I_{GREY}$}
            \State $M_c \leftarrow \{m_{cx}=0, m_{cy}=0\}$
            \State {$[r_I, c_I]$ = $SizeOf$($I_{GREY}$) }
            \If {$m_x<p$}
            \State $m_{cx}=p$
            \ElsIf{$m_x>c_I-p-1$}
            \State $m_{cx}=c_I-p-1$
            \Else
            \State $m_{cx}=m_x$
            \EndIf
            \If {$m_y<p$}
            \State $m_{cy}=p$
            \State $m_{cy}=r_I-p-1$
            \ElsIf{$m_y>r_I-p-1$}
            \State $m_{cy}=r_I-p-1$
            \Else
            \State $m_{cy}=m_y$
            \EndIf
            \State Return $M_c \leftarrow \{m_{cx}, m_{cy}\}$
            \EndFunction
        \end{algorithmic}
\end{algorithm}

\section{Experiment and result}
\label{sec:experiment}

\begin{figure}[htbp]
\centering
\begin{minipage}[t]{0.99\linewidth}
    \centering
    \includegraphics[width=1\columnwidth]{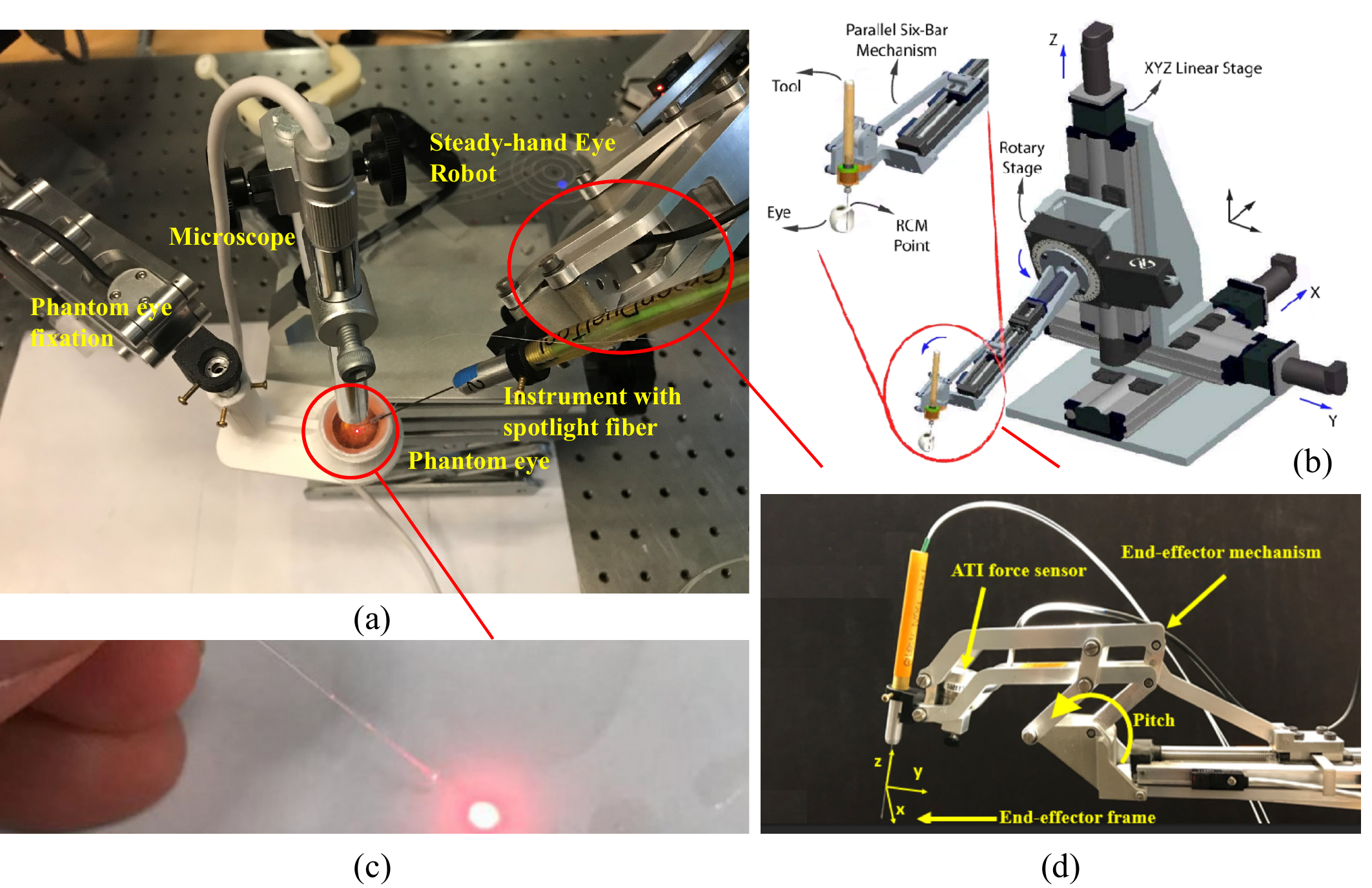}
\end{minipage}
\caption{(a) The overall experimental platform with SHER. (b) The detail structure of SHER. (c) The spotlight fiber with glue made lens in the lab. The fiber is placed on the surgical tool. (d) The end effector of SHER.}
\label{fig:fig_exp}
\end{figure}

\begin{figure*}[htbp]
\centering
\includegraphics[width=1\linewidth]{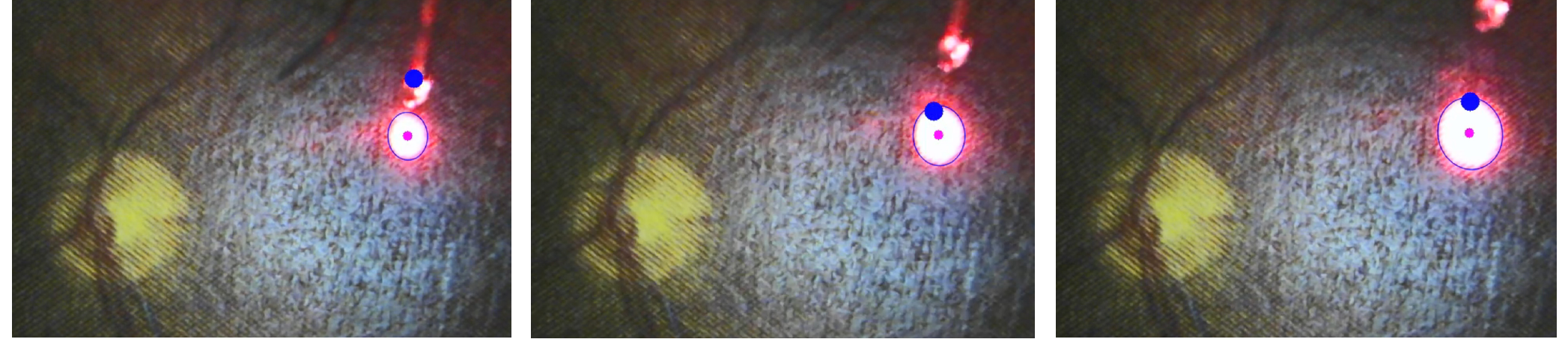}
\caption{The projection pattern changes with distance of the spotlight end-tip to surface. The distance change at a speed of 0.1 mm/s. The initial point for 0 mm distance is recorded when the fiber slightly touch the surface.}
\label{fig:dyn}
\end{figure*}

\begin{figure*}[htbp]
\centering
\includegraphics[width=1\linewidth]{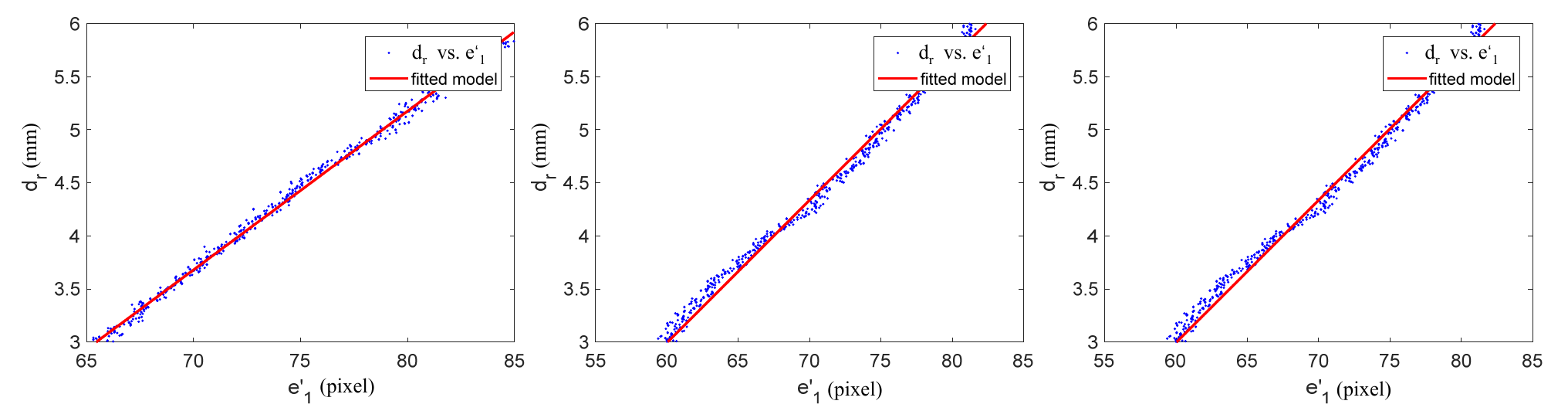}
\caption{Three trials with different location on the phantom plane. $e'_1$ denotes the  the short axis of the ellipse in microscope image. $d_r$ denotes the ground truth distance.}
\label{fig:fit_plane3}
\end{figure*}

\begin{figure*}[htbp]
\centering
\includegraphics[width=1\linewidth]{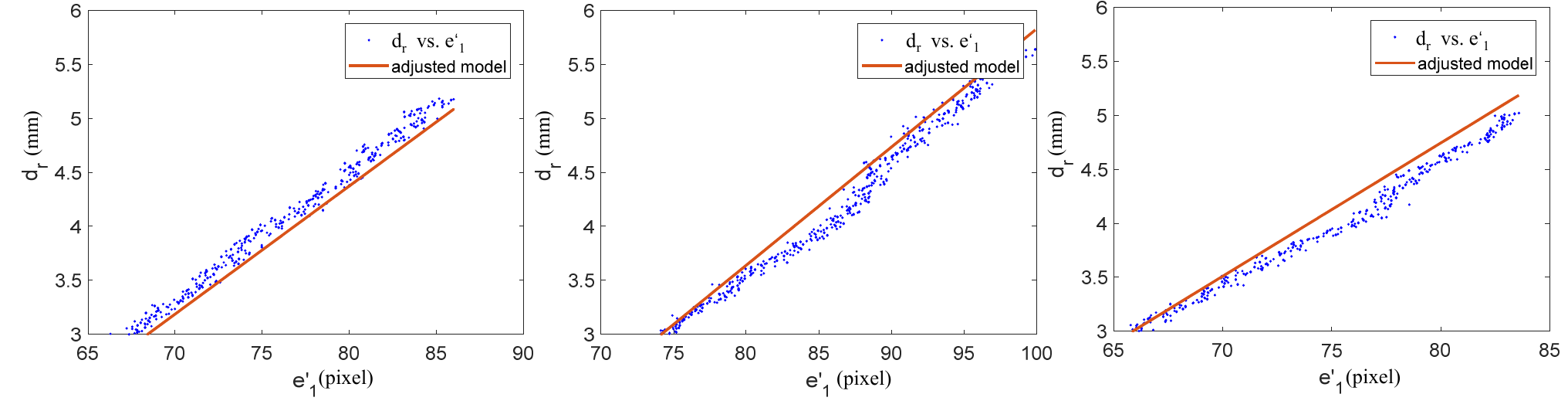}
\caption{Three trials with different location on the phantom eyeball. }
\label{fig:fit_sphere}
\end{figure*}

\begin{figure*}[htbp]
\centering
\includegraphics[width=1\linewidth]{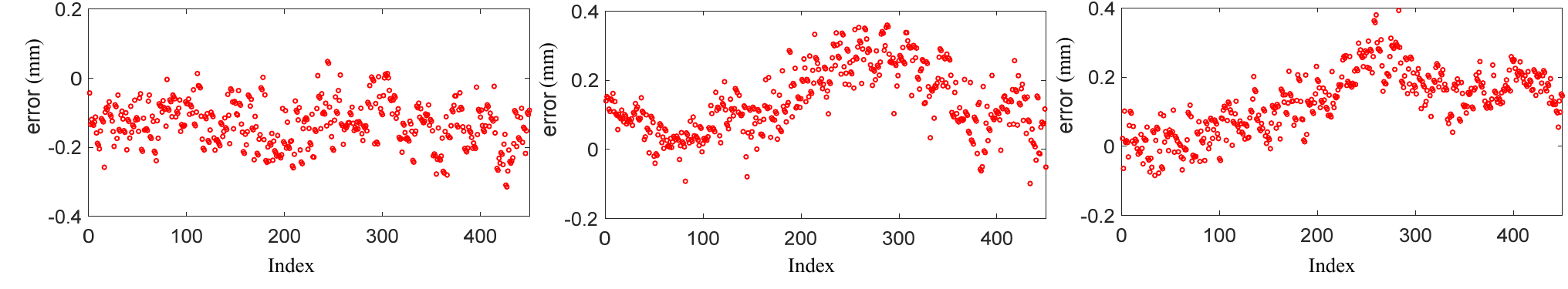}
\caption{The error performance for the three trials on eyeball phantom with adjusted model and ground truth distance. Index is the data point index.}
\label{fig:fit_error}
\end{figure*}

To verify the proposed method, we set up the experimental platform shown in Fig.~\ref{fig:fig_exp}(a). The phantom plane and phantom eye can be selected and fixed with phantom fixation. The human retina image is printed and attached on the phantom to make the background realistic. The surgical instrument with spotlight fiber is mounted on the SHER and directed at the phantom. The spotlight fiber is created by a normal 0.5 mm diameter light fiber with super glue to create a glue lens in the end-tip shown in Fig.~\ref{fig:fig_exp}(c). This simple but efficient method creates a suitable spotlight pattern size easily. SHER is a cooperatively controlled 5-DoF robot for microsurgical manipulation in eye surgery that helps surgeons reduce hand tremor and have more stable, smooth and precise manipulations, as shown in  Fig.~\ref{fig:fig_exp}(b)~\cite{ebrahimi2019adaptive,uneri2010new}. SHER is a velocity-controlled robot and the velocity of the motors of each joint is controlled by an embedded velocity controller (Galil 4088, Galil, 270 Technology Way, Rocklin, CA 95765). The cooperative control is realized by the 6-DoF ATI Nano17 force/torque sensor (ATI Industrial Automation, Apex, NC) mounted on the end-effector mechanism shown as Fig.~\ref{fig:fig_exp}(d), the force sensor can measure the 6 DoF force and toque applied on the end-effector.

We first perform the calibration experiment on the plane platform. The light fiber attached to the instrument is relocated to slightly touch the plane to sense the initial point. Afterward, the robot is constrained and controlled to move at a speed of 0.1 mm/s in the $Z$ direction of the end effector frame shown in Fig.~\ref{fig:fig_exp} (d). Thus the movement in $Z$ direction is recorded as the distance change of the spotlight end-tip to surface, shown as Fig.~\ref{fig:dyn}, which can be denoted as the ground truth value of the distance $d_r$. Based on Eq.~\ref{equa:basic7}, the value of $\delta k$ can be estimated based on the $e_1'$ calculated from the image and $d_r$. Three trials with different location on the phantom plane are tested. The fitting result is shown in Fig.~\ref{fig:fit_plane3}, and Table~\ref{tab:test1}. The average model is listed in Table~\ref{tab:test1} which represents the model for all data of three points. The linear fitting coefficient of determination $R^2$ is very close to 1 which means the data is fitted with the proposed model very well. The variation of the fitted model may come from the imperfect plane and the inaccuracy of the touch method for defining of initial point. The average model for the root mean square error (RSME) is 0.163 mm which demonstrates the performance of proposed method at a very slow speed 0.01 mm/s. Based on the known parameters of $\delta k$ and $b$ with the previous deduction, in the eyeball phantom, we only need to know $c$ then we could estimate the distance from the end-tip to sphere surface. To test the performance of the estimation error on the eyeball phantom, three points location inside the phantom eyeball is selected and test with the same procedure on the phantom plane. The adjusted model with the ground truth distance are shown in Fig.~\ref{fig:fit_sphere} and Table~\ref{tab:test2}, which we can find that $e_1'$ vs $d_r$ shows linear relationship very well. The error performance shown in Fig.~\ref{fig:fit_error} indicates that the maximum error is within 0.372 mm. The error may cause by the simplified mathematical model and the imperfect phantom sphere surface for the eyeball model.

\begin{table}[htbp]
\begin{center}
\caption{The three trials on the different location of phantom plane (in mm).}
\label{tab:test1}
\begin{tabular}{|c|c|c|c|c|c|c|}
 \hline
  &$\delta k$ & b & $R^2$ &RSME \\
  \hline
 point 1& 0.1496& -6.791& 0.9959 & 0.0548\\
 point 2& 0.1343&-5.069& 0.9909 &0.0974  \\
 point 3& 0.1347&-5.226&0.9961 & 0.0663\\
 average &  0.1321& -5.117 & 0.9703&0.1630 \\
 \hline
\end{tabular}
\end{center}
\end{table}

\begin{table}[htbp]
\begin{center}
\setlength{\tabcolsep}{2.5pt}
\caption{The three trials on the different location of phantom sphere (in mm).}
\label{tab:test2}
\begin{tabular}{|c|c|c|c|c|c|c|}
 \hline
  & c & standard deviation & mean absolute error & max error \\
  \hline
 point 1& 5.531& 0.064& 0.146 & 0.328\\
 point 2&  6.982&0.097& 0.140 &0.372  \\
 point 3& 4.584&0.088&0.160 & 0.396\\
 \hline
\end{tabular}
\end{center}
\end{table}

\begin{figure}[htbp]
\centering
\begin{minipage}[t]{1.0\linewidth}
    \centering
    \includegraphics[width=1\columnwidth]{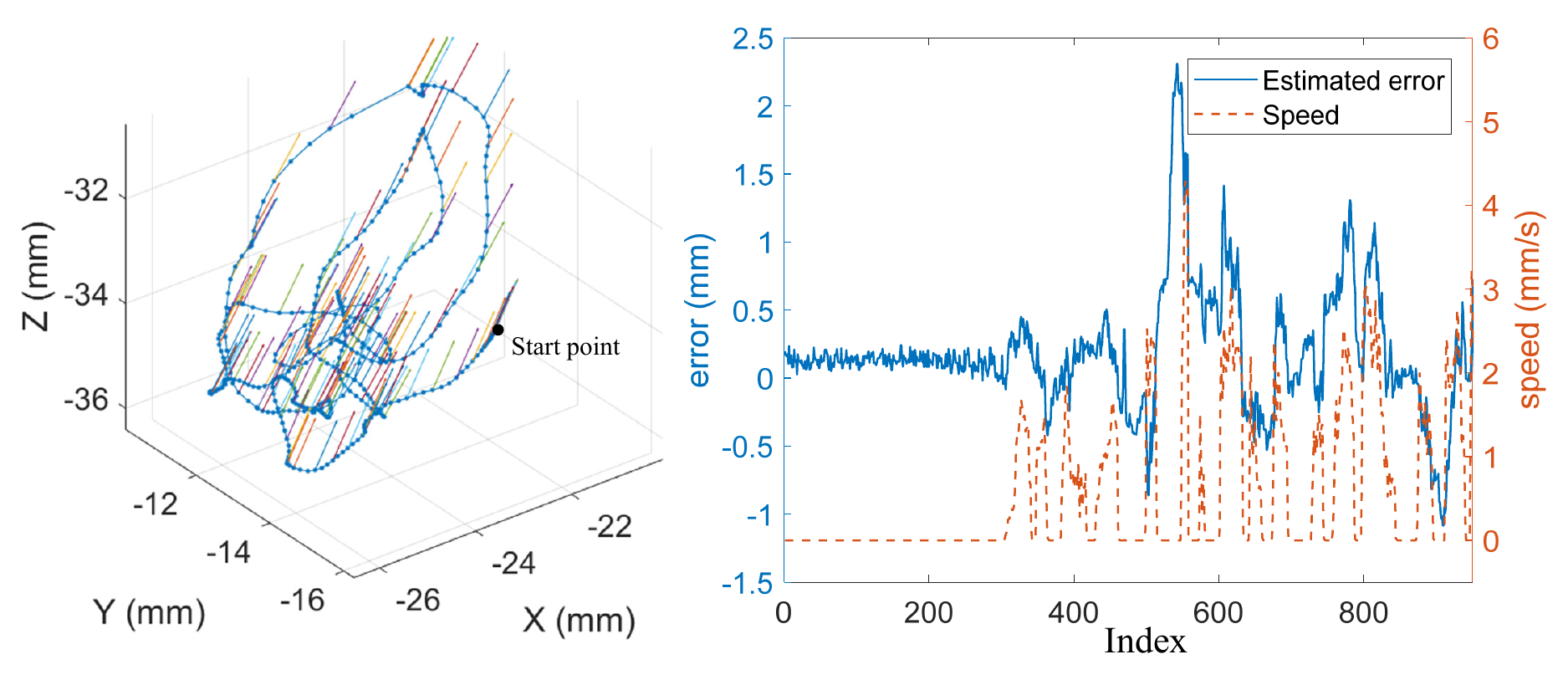}
        \parbox{6cm}{\small \hspace{0.9cm}(a)\hspace{3.6cm}(b)}
\end{minipage}
\caption{(a) The trajectory of spotlight pose of position in 3D space. The arrow denotes the pose of spotlight. The dot denotes the end-tip of spotlight. The blue line denotes the end-tip trajectory for the spotlight. (b) The speed of the spotlight end-tip and the estimated error for the distance between end-tip and phantom plane during motion.}
\label{fig:dyn_plane}
\end{figure}

In order to test the dynamic performance for the estimation of distance, we move the SHER with cooperative control mode, during which the robot is controlled by the operator's hand. The instrument's trajectory is calculated by the forward kinematics of the robot with updating at 200 Hz of position information from joint sensors, which is shown as Fig.~\ref{fig:dyn_plane}(a). The projection of spotlight light on the phantom is recorded and calculated by the USB microscope camera with 10 Hz. The distance between spotlight end-tip and the plane phantom is calculated by the position and pose of the spotlight mounted on the instrument with the known plane formulation in space. The plane formulation is known by using the robot to touch the three points on the phantom plane. The relationship between the estimated distance error and speed of the spotlight end-tip could be found in Fig.~\ref{fig:dyn_plane}(b). It shows  that the estimated error has a positive correlation with the speed, with higher  end-tip speeds causing larger error. This is due to the faster speed that will cause the image tail of the spotlight projection, which makes the inaccuracy of the estimation of $e_1'$. This could be potentially improved by using a faster and more sensitive camera. In applications with a combination of OCT imaging or keep the safe distance between the instrument tip to retinal surface, 0.5 mm would be an acceptable error. Based on the observation of Fig.~\ref{fig:dyn_plane}(b), to restrict the error into 0.5 mm the speed of the end-tip of spotlight should be controlled within 1.5 mm/s in the test scenario. 

\section{Conclusion and future work}
\label{sec:conclusion}
In this paper, we demonstrate a novel method for 3D guidance of an instrument using the projection of a spotlight in a single microscope image. The spotlight projection mechanism is analyzed and modeled on both a plane and a spherical surface. We then design and apply an algorithm to track and segment the spotlight projection pattern. Here the SHER with the spotlight setup, is used to verify the proposed idea. The results demonstrates that the static performance of the proposed method has a RSME of 0.163 mm on the plane phantom and maximum error of 0.372 mm on the spherical eyeball phantom. The dynamic error performance is dependent on the speed of the spotlight end-tip motion. To restrict the error within 0.5 mm, the speed of the end-tip of spotlight should be limited to 1.5 mm/s as evidenced in the test scenario. In next steps, we plan to validate the proposed method by performing visual navigation tasks relevant to retinal surgery and to explore potential benefits and opportunities for further refinement.

%
%

%

\balance

\bibliographystyle{IEEEtran}
{\scriptsize
\bibliography{MyCollection}
}
\end{document}